\documentclass{article}
\usepackage{spconf,amsmath,graphicx,hyperref}
\usepackage{amsfonts}
\usepackage{algorithmic}
\usepackage{algorithm}
\usepackage{array}
\usepackage[table]{xcolor}
\usepackage{textcomp}
\usepackage{stfloats}
\usepackage{url}
\usepackage{verbatim}
\usepackage{cite}
\usepackage{booktabs} 
\usepackage{multirow} 
\usepackage{amssymb}
\usepackage{dsfont}
\usepackage{caption}
\usepackage{pifont}
\usepackage{enumitem}
\usepackage{subcaption}
\setlist[itemize]{noitemsep, topsep=0pt}

\title{DAM: Dual Active Learning with Multimodal Foundation Model for Source-Free Domain Adaptation}
\name{Xi Chen$^{1}$ \qquad Hongxun Yao$^{1}$\sthanks{Corresponding author.} \qquad Zhaopan Xu$^{1}$ \qquad Kui Jiang$^{1}$}
  
  \address{$^{1}$ Harbin Institute of Technology}
\definecolor{deepblue}{rgb}{0.2,0.4,0.8}

\newcommand{\ie}{\text{i}.\text{e}.{}}
\newcommand{\eg}{\text{e}.\text{g}.{}}

\newcommand{\fullname}{\underline{D}ual \underline{A}ctive learning with \underline{M}ultimodal (DAM) foundation model}
\newcommand{\shortname}{DAM}
\newcommand{\selectorname}{Dual-Focused active Supervision}
\newcommand{\shortselectorname}{DFS}
\newcommand{\distname}{Alternative Distillation}
\newcommand{\shortdisname}{ADL}

\begin{document}
\maketitle
\begin{abstract}
Source-free active domain adaptation (SFADA) enhances knowledge transfer from a source model to an unlabeled target domain using limited manual labels selected via active learning. While recent domain adaptation studies have introduced Vision-and-Language (ViL) models to improve pseudo-label quality or feature alignment, they often treat ViL-based and data supervision as separate sources, lacking effective fusion. To overcome this limitation, we propose \fullname{}, a novel framework that integrates multimodal supervision from a ViL model to complement sparse human annotations, thereby forming a dual supervisory signal. \shortname{} initializes stable ViL-guided targets and employs a bidirectional distillation mechanism to foster mutual knowledge exchange between the target model and the dual supervisions during iterative adaptation. Extensive experiments demonstrate that \shortname{} consistently outperforms existing methods and sets a new state-of-the-art across multiple SFADA benchmarks and active learning strategies.
\end{abstract}


\section{Introduction}
Deep neural networks often exhibit limited generalization to unseen domains due to distributional shift between training (source) and target data. This challenge is prevalent in applications such as autonomous driving, remote sensing, and medical imaging. Unsupervised domain adaptation (UDA)~\cite{DAPL,RADA-prompt} addresses this shift by aligning feature distributions across domains; however, its dependence on source data introduces privacy concerns~\cite {AdaMix} and logistical challenges. Source-free domain adaptation (SFDA)~\cite{SHOT,A2Net,sfda_survey} offers a more practical alternative by adapting a pre-trained source model without access to the original data, though this constraint makes adaptation more challenging. Allowing minimal labeling effort on the target data can yield substantial gains further for SFDA, motivating the study of source-free active domain adaptation (SFADA)~\cite{MHPL,SQAdapt,BIAS}.

Meanwhile, Vision-and-Language (ViL) foundation models, \eg, CLIP~\cite{clip} and ALIGN~\cite{ALIGN} have demonstrated remarkable zero-shot generalization and potential for downstream knowledge transfer. Their rich multimodal representations are promising for bridging domain gaps, especially in low-data regimes, \eg, via prompt tuning~\cite{Coop} or adapter modules~\cite{Tip-Adapter}. 
This prompts a key research question: \textit{Under a fixed labeling budget, how can we best leverage ViL models to improve the adaptation of source-domain-pretrained models to a target domain?}
Although several studies~\cite{DAPL,DIFO,RADA-prompt} have primarily incorporated ViL models for enhanced pseudo-labeling or feature alignment, they often treat the ViL model and data as separate sources of pseudo-supervision. For instance, a straightforward combination that merely integrates active samples~\cite{MHPL} with a ViL-based method like DIFO~\cite{DIFO} yields only marginal gains (+1.1\%/0.4\% on Office-Home/VisDA-C datasets)--significantly lower than the +7.5\%/8.9\% improvement achieved by methods that rely solely on data pseudo-supervision without ViL components~\cite{MHPL}. This result underscores that effectively leveraging ViL models requires more sophisticated interaction mechanisms beyond simple combination.

To address this combination gap, we propose \fullname{}. As shown in Fig.~\ref{fig:compare}, our approach augments the target model with auxiliary supervision from a ViL model throughout adaptation.
However, since the ViL model is source-agnostic and lacks task-specific knowledge, its representations must be carefully activated to provide reliable guidance.
To this end, beyond conventional active sample supervision~\cite{MHPL,ELPT,CoreSet1}, we design a more stable ViL-based signal by prompt-tuning with those active samples, forming a \selectorname{} (\shortselectorname).
Furthermore, to fully exploit both supervisions, we introduce an \distname{} (\shortdisname) strategy that facilitates bidirectional knowledge transfer between the supervisions and the target model throughout adaptation. This enables mutual enhancement by exchanging the ViL model's generalization capacity with the target-specific representations learned by the target model, thereby improving adaptation performance for both.

In summary, our main contributions are as follows: 
(1) We propose \shortname{}, a novel framework that integrates ViL-based supervision with active learning to construct a dual supervisory signal, significantly enhancing SFADA under a limited budget.
(2) We design a \shortselectorname{} mechanism to activate task-aware ViL supervision using selected target samples, and an \shortdisname{} protocol to enable mutual knowledge exchange between the target model and the dual supervisions.
(3) Extensive experiments show that \shortname{} sets new state-of-the-art results across multiple benchmarks and active learning strategies.

\section{METHODOLOGY}
\label{sec:format}
\begin{figure*}[th]
    \begin{minipage}[t]{.25\textwidth}
        \centering
    \centerline{\includegraphics[width=.9\textwidth]{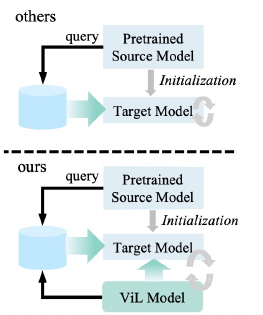}}
    \subcaption{}
    \label{fig:compare}
    \end{minipage}
    \vrule
    \hspace{10pt}
    \begin{minipage}[t]{.65\textwidth}
    \centering
    \centerline{\includegraphics[width=.9\textwidth]{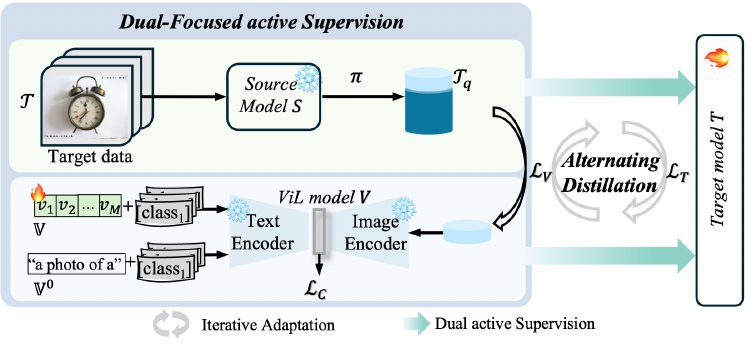}}
    \subcaption{}
    \label{fig:frame}
    \end{minipage}
    \caption{(a): Comparison: conventional SFADA methods~\cite{MHPL,SQAdapt,ELPT} vs. our approach with ViL model supervision. (b): Overview of \shortname{}. The target samples $\mathcal T$ are first sent to source model $S$, selecting $\mathcal T_q$ with strategy $\pi$. ViL model is tuned for auxiliary active supervision, forming \textit{Dual-focused active Supervision}. Moreover, \textit{Alternating Distillation} enables bidirectional knowledge transfer for target model adaptation.}
\end{figure*}
In the context of SFADA, 
a source model $S$ is first pretrained on a labeled source dataset $\{(x_i,y_i)\}_{i=1}^{m}$ spanning $C$ classes.
The \textbf{goal} of SFADA is to select a subset of labeled samples $\mathcal{T}_q$ from unlabeled target data $\mathcal{T}=\{x'_i\}_{i=1}^{n}$ within a labeling budget $B$, using an active query strategy $\pi$, such that the target risk of the target model $T$ on $\mathcal T$ is minimized. The target model $T$ is initialized with source parameters $T=S$, and adaptation is performed without access to the source data $\mathcal{S}$. Following~\cite{MHPL}, we adopt a one-shot query strategy, where all queried samples $\mathcal{T}_q$ are selected according to raw source model $S$ under the constraint $|\mathcal{T}_q|=\rho|\mathcal{T}|$, with $\rho \in (0,1)$. The labels of these samples are obtained from an oracle and subsequently used for model adaptation.

In this work, we incorporate a ViL model $V$ as an auxiliary source of prior knowledge for $T$. Let $f^I$ represent the image features extracted by the frozen image encoder of $V$.
The corresponding text embeddings are derived from a set of learnable category prompts $\{\mathbb{V}_k\}_{k=1}^C$, where each prompt $\mathbb{V}_k$ is structured as $\{v_1, v_2, \cdots, v_m, c_k\}$. These prompts are initialized using the template $\mathbb{V}^0_k:\text{``a photo of a }[c_k]\text{''}$ and are fed into the frozen text encoder to obtain text embeddings $\{\mathbf{w}_k\}_{k=1}^C$ such that $\mathbf{w}_k = \mathrm{Enc}_t(\mathbb{V}_k)$.
The prediction probability for class $k$ given an input image $x$ is computed as:
\begin{equation}
    p_V^{(k)}(x)= \frac{\exp(\text{sim}(f^I,\mathbf w_k)/\tau)}{\sum_{i=1}^C\exp(\text{sim}(f^I,\mathbf w_i)/\tau)},
    \label{eq:align}
\end{equation}
where $\text{sim}(\cdot,\cdot)$ denotes cosine similarity, and $\tau$ is a learned temperature parameter within ViL model. Throughout the adaptation, both the text and image encoders of the ViL model are frozen; only the category prompts are optimized to specialize the model to the target domain. 

The overall architecture is illustrated in Fig.~\ref{fig:frame}. 

\subsection{Dual-Focused Active Supervision (\shortselectorname{})}
In addition to conventional data supervision from the labeled subset $\mathcal T_q$ under different sampling strategies $\pi$, we incorporate supervisory signals from the ViL model using the same set. As the query ratio $\rho$ is small, to effectively adapt the ViL model to the target domain and prevent overfitting, we follow a supervised prompt learning strategy~\cite{Co_Coop} to preserve the model's foundational generalization capabilities.
This design choice is grounded in the ViL model's inherent capacity for robust multimodal alignment.
Specially, after the one-shot active sample query for $\mathcal T_q$, the context vectors $\mathbb{V}$ are optimized by minimizing the following \textbf{C}ross-\textbf{E}ntropy loss $\mathcal L_{ce}$ and regularization loss $\mathcal L_{kg}$: 
\begin{align}
\mathcal L_{C}&= \mathcal L_{ce} + \mathcal L_{kg} \notag\\
     &= \mathbb E_{x\in\mathcal T_q} \text{CE}\big(\mathbf e_y\,\|\, p_V(x)\big)
     + \frac{1}{C}\sum_{k=1}^{C} ||\mathbf w_k-\mathbf w_k^0||_2^2,
    \label{loss:ce_v}
\end{align}
where $\mathbf e_y$ denotes the one-hot vector of the ground-truth label $y$ (\ie, $\mathbf e_y\in\{0,1\}^C$ and $\mathbf e_y(k)=1$ if $k=y$), and $\mathcal L_{kg}$ regularize the textual embeddings $\mathbf w_k$ by anchoring them to their original general-purpose counterparts $\mathbf w^0_k=\text{Enc}_t(\mathbb V_k^0) $ to maintain its generalization.

\subsection{Alternating Distillation (ADL)}
To further use the dual-path supervision signals for target model adaptation to unlabeled target data, we propose an alternating distillation mechanism that aims to combine the task-specific discriminative knowledge from the target model and labeled samples $\mathcal T_q$ with the generalizable cross-modal knowledge from the ViL model. 

Specifically, the data supervision is utilized in this distillation process via a direct weighting scheme that emphasizes reliable samples $\mathcal T_q$ with oracle annotations while down-weighting samples with uncertain pseudo-labels. The unified mutual teacher framework provides supervision in the following manner: for a student model \(A\in\{T,V\}\) taught by teacher model \(B\in\{V,T\}\), we define the teacher's output distribution and sample weights $\mathcal W(\cdot)$ as:
\begin{align}
\big(\tilde p_{A\leftarrow B}, \mathcal W\big)=
\begin{cases}
\big(e_{y},\beta_q\big), x\in \mathcal T_q\ \text{with label }y,\\
\big(\mathrm{F}(B(x)/\tau_B),\beta\big), x\in \mathcal T\setminus \mathcal T_q,
\end{cases}    
\label{eq:teacher}
\end{align}
where $\mathrm{F}(\cdot)$ is $\text{Softmax}(\cdot)$ operation,  \(\beta_q \ge \beta>0\) are hyperparameters controlling the weight contribution. The student's prediction is given by
\(p_A^{\tau_A}(x)=\mathrm{F}(A(x)/\tau_A)\).
Note that hard pseudo-labeling emerges as a special case when \(\tau_B\!\to\!0\), which reduces the teacher's distribution to a one-hot vector.

\subsubsection{Distillation from $T\rightarrow V$}
During each epoch, we select the top-$N$ most confident samples per class from the target model's predictions: $X_{\text{top}}=\bigcup \limits_{y\in\mathcal{Y}_t}$ top-$N$ by $\max_k T^{(k)}(x)$ within class $y$ and enable task-specific knowledge distillation to the ViL model through their associated soft labels $T(x)$ with $\tau_V=\tau_T=1$ by:
\begin{equation}
    \mathcal L_{dist}^{V\leftarrow T} =
        -\mathbb{E}_{x \in {X}_{\text{top}}\cup\mathcal T_q}\mathcal W(x)\text{CE}\big(\tilde p_{V\leftarrow T}(x)||p_V^{\tau_V}(x)\big).
\end{equation}

\subsubsection{Distillation from $V\rightarrow T$}
Conversely, since the ViL model outputs multimodal cosine similarities as predictions, the hard labels produced by the ViL model are used as pseudo-labels for unlabeled samples $\mathcal T \setminus \mathcal T_q$ to supervise the adaptation of the target model. The distillation loss for the target model is formulated as:
\begin{equation}
    \mathcal L_{dist}^{T\leftarrow V}= -\mathbb E_{x\in{\mathcal{T}}}\mathcal W(x)\text{CE}\big(\tilde p_{T\leftarrow V}(x)||p_T^{\tau_T}(x)\big),
\end{equation}
where $\tau_T=1$ and $\tau_V\rightarrow0$.

Following prior works~\cite{SHOT,DIFO,GRAPE}, we incorporate entropy minimization and diversity regularization loss terms to guarantee the unambiguous and balanced classes:
\begin{align}
     \mathcal L_{ent} &=\mathbb E_{x\in \mathcal T}\text{CE}\Big(p_T(x)\,\|\,p_T(x)\Big),\\
    \mathcal{L}_{div} &= \sum_{k=1}^{C}\hat p_k\log(\hat p_k) =\text {KL}(\hat p,\frac{1}{C}\mathds{1}(C)) - \log C, 
\end{align}
where $\hat p = \mathbb{E}_{x\in \hat{X}_t}[p_T(x)]$ represents the average output embedding for the whole target dataset, KL is the KL distance, and $\mathds{1}(C)$ is a vector of ones with dimension $C$.

The overall training losses are then defined as:
\begin{equation}
    \mathcal L_T = \mathcal L_{\mathrm{dist}}^{T\leftarrow V}+ \mathcal L_{ent} + \mathcal{L}_{div},
    \label{loss:dis_t}
\end{equation}
\begin{equation}
    \mathcal L_V=\mathcal L_{\mathrm{dist}}^{V\leftarrow T}+\mathcal L_{kg}.
    \label{loss:dis_v}
\end{equation}
After training, the ViL model is discarded, and only the target model is used for inference.

\section{EXPERIMENT}
\label{sec:exp}
\subsection{Datasets} We evaluate our method on three benchmarks: 
(1) Office~\cite{office-31}: A small-scale dataset across \textbf{A}mazon, \textbf{D}slr, and \textbf{W}ebcam domains; (2) Office-Home~\cite{office-home}: A medium-scale dataset that is distributed across four domains: \textbf{A}rt, \textbf{C}lipart, \textbf{P}roduct and \textbf{R}eal-World; (3) VisDA-C~\cite{Visda-c}: A large-scale dataset designed for \textbf{S}ynthetic-to-\textbf{R}eal adaptation.

\subsection{Baselines}
Our method is evaluated against comprehensive baselines, spanning five categories (1) Source(model)-only and zero-shot CLIP with ViT-B/32 backbone (ZS-C-B32), where we use the standard prompt ``a photo of a [class]``; (2) Unsupervised DA (UDA) methods; (3) Source-free DA (SFDA) approaches; (4) Active DA (ADA) methods; and (5) Source-free active DA (SFADA) techniques. 

\subsection{Implementation}
Our implementation for training source and models follows the established training pipeline of prior works~\cite{SHOT,GRAPE,MHPL}. Specifically, ResNet-50~\cite{Resnet} is used for Office and Office-Home datasets, while ResNet-101~\cite{Resnet} is employed for VisDA-C. For the ViL model component, we use the CLIP model~\cite{clip} with a ViT-B/32 image encoder~\cite{vit} across all experiments. The hyperparameters in Eq.~\ref{eq:teacher} are set as $\beta_q=3$ and $\beta=0.3$.
The context vector length $m$ of prompts is set to 16, and the $X_{top}$ set is constructed with $N=16$. 
In \shortselectorname{}, the prompts are optimized using SGD with a base learning rate of $2 \times 10^{-3}$, incorporating a warmup strategy starting at $1 \times 10^{-5}$ for the first epoch and followed by cosine annealing over 50 epochs. In \shortdisname{}, prompt learning rate is maintained at $2 \times 10^{-3}$.
All experiments were run on a single NVIDIA RTX 4090 GPU and conducted with a labeling budget of $\rho=5\%$ unless otherwise specified. 

\begin{table}[!t]
\centering
\caption{Performance (\%) on three datasets  with a 5\% labeling budget. SF: source-data-free; ViL: use ViL models; 'OH': Office-Home dataset. $\dag$: Reimplemented using official code with source models identical to ours.}
\resizebox{0.98\linewidth}{!}{
\begin{tabular}{l|c|l|c|c|c|c}
   \toprule
     Categories&SF&Method& ViL &Office&OH &VisDA-C\\
    \midrule
    \multirow{2}*{Zero-shot} &\multirow{2}*{\checkmark}& Source-only&\ding{55} &77.2& 58.7 &46.9 \\
    &&ZS-C-B32~\cite{clip} &\checkmark&79.5&72.2 &87.2 \\
    \midrule
    \multirow{2}*{UDA} &\multirow{2}*{\ding{55}}&
    RADA-prompt~\cite{RADA-prompt} &\ding{55}&91.6&72.9&90.5 \\
    &&DAPL~\cite{DAPL} &\checkmark &- &74.5 & 86.9\\
    \midrule
    \multirow{5}*{SFDA}&\multirow{5}*{\checkmark} & 
    CPGA~\cite{CPGA}&\ding{55}&89.9&71.6&86.0\\ 
    &&SHOT~\cite{SHOT}&\ding{55}&88.6&71.8&82.4\\
    &&A2Net~\cite{A2Net}&\ding{55} &90.1&72.8&84.3\\
    &&GRAPE~\cite{GRAPE}&\ding{55} &91.6&75.4&89.2 \\
    &&DIFO-C-B32\dag ~\cite{DIFO}& \checkmark&92.6&76.3&90.1\\

    \midrule
    \multirow{4}*{ADA} &\multirow{4}*{\ding{55}}
    &CLUE~\cite{CLUE} &\ding{55}& 89.5&72.5&-\\
    &&EADA~\cite{EADA} &\ding{55}&93.2&76.7&88.3\\
    &&DUC~\cite{DUC} &\ding{55}&-&78.0&88.9\\
    &&CA-ADA~\cite{CA-ADA}&\ding{55}&-&79.3&-\\
    \midrule
    \multirow{8}*{SFADA}&\multirow{8}*{\checkmark}
    &SQAdapt~\cite{SQAdapt} &\ding{55}&-&77.8&91.3\\
    \cline{3-7}
    &&ELPT~\cite{ELPT} &\ding{55} &92.8& 77.0 &89.2\\
    &&+\shortname{}     &\checkmark&\textbf{93.5}&\textbf{81.9}&\textbf{91.9} \\
    \cline{3-7}
    &&CoreSet~\cite{CoreSet1} &\ding{55} &90.3& 75.2&85.9 \\
    && +BIAS~\cite{BIAS} & \ding{55} &91.0&76.8&88.2\\
    &&+\shortname{} &\checkmark &\textbf{93.4}& \textbf{83.9}&\textbf{92.0} \\
    \cline{3-7}
    &&MHPL~\cite{MHPL} &\ding{55} &93.2&79.3&91.3\\
    &&+BIAS~\cite{BIAS}&\ding{55}&93.6& 80.3&91.1\\
    &&+\shortname{}&\checkmark &\textbf{94.1}&\textbf{82.0}&\textbf{92.0}\\
    \bottomrule
    \end{tabular}}
\label{tab:office_home}
\end{table}

\subsection{Main Results}
Table~\ref{tab:office_home} demonstrates the superior performance of \shortname{} across multiple benchmarks. Comparative analysis with BIAS~\cite{BIAS}--a plug-and-play framework that aligns distributions of inactive and active samples--reveals consistent improvements across all datasets and base methods. Specifically, \shortname{} combined with CoreSet and MHPL outperforms their BIAS-integrated counterparts, achieving accuracy gains of 2.4\%/7.1\%/3.8\% and 0.5\%/1.7\%/0.9\% over CoreSet+BIAS and MHPL+BIAS on Office, Office-Home, and VisDA-C, respectively. Notably, \shortname{} surpasses both UDA and ADA methods that require access to source data, confirming the efficacy of leveraging ViL model supervision.



\begin{figure}[!t]
\begin{minipage}[b]{.48\linewidth}
    \centering
\centerline{\includegraphics[width=1.\linewidth]{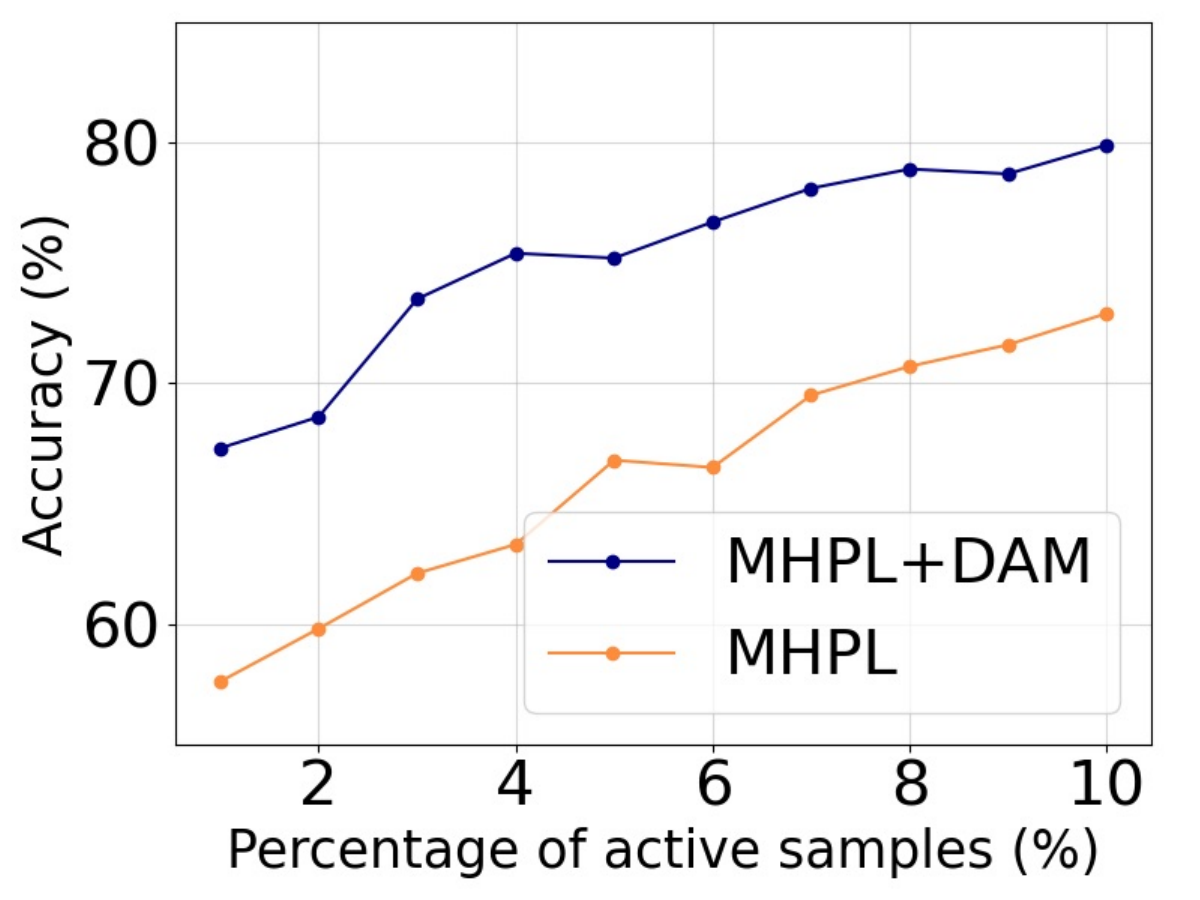}}
\subcaption{$A\rightarrow C$}
\end{minipage}
\begin{minipage}[b]{.48\linewidth}
    \centering
\centerline{\includegraphics[width=1.\linewidth]{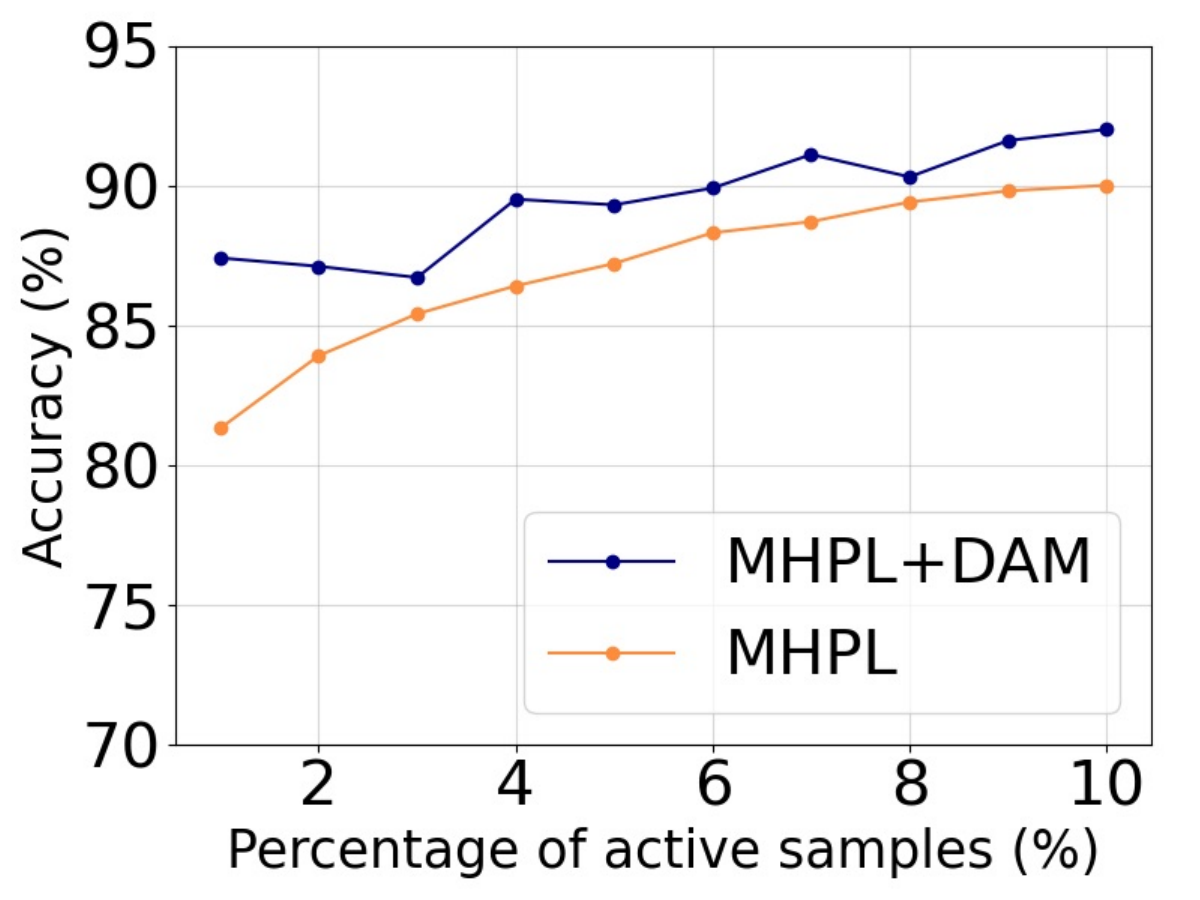}}
\subcaption{$C\rightarrow P$}
\end{minipage}
\hfill
    \caption{ Ablation study of model performance with \shortname{} across different labeling quotas for active learning on the transfer task (a) $A\rightarrow C$ and (b) $C\rightarrow P$ of Office-Home dataset.}
    \label{fig:ratio_abl}
\end{figure}

\begin{table}[!t]
\caption{Ablation study of $\mathcal L_C$ and $\mathcal L_V$ on two transfer tasks in the Office-Home dataset.}
\centering
    \setlength{\tabcolsep}{10pt}
\resizebox{0.81\linewidth}{!}{
    \begin{tabular}{l|*{2}{c}|*{2}{c}}
   \toprule
   Method&$\mathcal L_{C}$ &$\mathcal L_{V}$&$A\rightarrow C$ & $C \rightarrow  P$ \\
   \midrule
    Source+CLIP&&&69.8 &85.2 \\
   \textit{w/} $\mathcal L_C$&\checkmark&&72.7& 89.3\\
    \textit{w/} $\mathcal L_V$&&\checkmark& 70.5&85.8\\
    \shortname&\checkmark &\checkmark &\textbf{75.2}&\textbf{89.3}\\
    \bottomrule
    \end{tabular}}
\label{tab:abl_loss}
\end{table}

\subsection{Ablation Analysis}
Our analysis is based on MHPL+\shortname{} framework. 

\noindent\textbf{Loss Ablation.} 
To evaluate the individual contributions of the loss components in \shortname{}, we conduct an ablation study on the two losses applied to the ViL model. The results, presented in Table~\ref{tab:abl_loss}, use the direct combination of 'Source+CLIP' as the baseline. Our findings demonstrate the consistent and significant importance of each loss component for the overall performance.

\noindent\textbf{The Ratio $\rho$ of Active Samples.}
To investigate the impact of the active sample ratio $\rho$, we conduct experiments on the $A \rightarrow C$ and $C \rightarrow P$ transfer tasks from the Office-Home dataset. As shown in Fig.~\ref{fig:ratio_abl}, the performance of MHPL enhanced with \shortname{} consistently and significantly surpasses that of the base MHPL across all tested ratios. Notably, even under limited labeling budgets (i.e., small values of $\rho$), \shortname{} still delivers substantial performance improvements.

\begin{table}[!t]
    \centering
    \caption{Training resource comparison: parameter count and GPU memory usage with (\textit{w/}) and without (\textit{w/o}) the ViL model.}
    \resizebox{0.9\linewidth}{!}{
    \begin{tabular}{l|cc}
    \toprule
    \textit{ w/} ViL&Trainable Params[M]&Peak GPU Memory[GB] \\
     \midrule
        \ding{55} &24.041 &5.43\\
        \checkmark (ours) & 24.043(+0.002) &5.91(+0.48)\\
    \bottomrule
    \end{tabular}}
    \label{tab:cost}
\end{table}
\noindent\textbf{Computational Cost.}
Table~\ref{tab:cost} summarizes the computational costs during training. Our \shortname{} framework employs a prompt learning technique, adding merely 0.002M trainable parameters during training and a modest 0.48 GB (8.8\%) increase in GPU memory. Notably, the ViL module contributes zero additional FLOPs during inference as it is deactivated post-training, and only the target model is used. This optimized design preserves real-time performance while enhancing model capabilities.


\section{Conclusion and Future Work}
\noindent\textbf{Conclusion.} We propose \shortname, a novel plug-and-play framework for SFADA that enhances and improves supervision by combining sparse human annotations with rich multimodal knowledge from a ViL model via the \textit{\selectorname{}}. The adaptation process is optimized using \textit{\distname{}} between the target model and supervisions, which integrates task-specific priors and generalization capabilities. 
Extensive experiments on multiple benchmarks validate the effectiveness of our method's consistent and substantial improvements over SOTAs. 

\noindent\textbf{Future Work.} 
\shortname{} is compatible with various active sampling strategies, yielding different performance outcomes. A promising direction for future work is the exploration of optimal sampling methods within this collaborative architecture. As the present study focuses on ViL activation, we leave the investigation of sampling strategies for future research.

\vfill\pagebreak

{
\small
\bibliographystyle{IEEEbib}
\bibliography{arxiv}
}
\end{document}